\documentclass[conference]{IEEEtran}
\IEEEoverridecommandlockouts
\usepackage{cite}
\usepackage{amsmath,amssymb,amsfonts}
\usepackage{algorithmic}
\usepackage{graphicx}
\usepackage{textcomp}
\usepackage{xcolor}
\usepackage{pgfplots}
\usepackage{tikz}
\usepackage[autostyle]{csquotes}
\definecolor{c1}{RGB}{0,113,188}
\definecolor{c2}{RGB}{217,82,24}
\definecolor{c3}{RGB}{56,118,29}
\definecolor{c4}{RGB}{116,27,71}
\def\BibTeX{{\rm B\kern-.05em{\sc i\kern-.025em b}\kern-.08em
T\kern-.1667em\lower.7ex\hbox{E}\kern-.125emX}}

\begin{document}

\title{SteLLA: A Structured Grading System Using LLMs with RAG\\
}

\author{\IEEEauthorblockN{Hefei Qiu\IEEEauthorrefmark{1},
Brian White\IEEEauthorrefmark{2},
Ashley Ding\IEEEauthorrefmark{3},
Reinaldo Costa\IEEEauthorrefmark{2},
Ali Hachem\IEEEauthorrefmark{2},
Wei Ding\IEEEauthorrefmark{2} and
Ping Chen\IEEEauthorrefmark{2}}

\IEEEauthorblockA{\IEEEauthorrefmark{1}Department of Computer Science\\ Fitchburg State University, 160 Pearl Street, Fitchburg, MA 01420-2697\\
\IEEEauthorrefmark{2}Department of Computer Science\\
University of Massachusetts Boston,
100 Morrissey Blvd, Boston, MA 02125\\ Email: \{brian.white, reinaldo.costa001, ali.hachem002, wei.ding, ping.chen\}@umb.edu}
\IEEEauthorblockA{\IEEEauthorrefmark{2}Chantilly High School\\
4201 Stringfellow Rd, Chantilly, VA 20151\\
Email: ashley.ding6@gmail.com }}

\maketitle

\begin{abstract}
Large Language Models (LLMs) have shown strong general capabilities in many applications. However, how to make them reliable tools for some specific tasks such as automated short answer grading (ASAG) remains a challenge. We present SteLLA (\underline{St}ructured Grading System Using \underline{LL}Ms with R\underline{A}G) in which a) Retrieval Augmented Generation (RAG) approach is used to empower LLMs specifically on the ASAG task by extracting structured information from the highly relevant and reliable external knowledge based on the instructor-provided reference answer and rubric, b) an LLM performs a structured and question-answering-based evaluation of student answers to provide analytical grades and feedback. A real-world dataset which contains students' answers in an exam was collected from a college-level Biology course. Experiments show that our proposed system can achieve substantial agreement with the human grader while providing break-down grades and feedback on all the knowledge points examined in the problem. A qualitative and error analysis of the feedback generated by GPT4 shows that GPT4 is good at capturing facts while may prone to inferring too much implication from the given text in the grading task which provides insights into the usage of LLMs in the ASAG system.

%Open-ended problems such as short answer questions have been an effective way in learning assessment. But grading on such problems is very time-consuming which hinders the usage of them. In creating an automated short answer grading (ASAG) system, some key challenges include how to make one system applicable to variant subjects, how to improve the explainability, and how to increase users’ trust in using it. To address these challenges, we design an ASAG assistant which a) employs Large Language Models (LLMs) to make it be adaptable to any subject, b) applies clustering to select few shots and prompting to improve the performance of LLMs, c) uses Question Generation (QG) and Question Answering (QA) techniques to extract structured information from text answers to perform text comparison which also makes the grading more explainable, d) incorporates human in the loop to further increase the performance and users’ trust. The experiment results on a dataset from a college-level introductory Biology course exam show that our system saves a large percentage of the instructor’s time on grading while reaching close-to-human performance. The feedback produced by the system also serves great help to both instructors and learners. (to be updated)
\end{abstract}

\begin{IEEEkeywords}
LLM-based ASAG system, RAG, QA-based evaluation, structured evaluation
\end{IEEEkeywords}

\section{Introduction}
Assessment plays an important role in the teaching and learning process. It usually includes closed-ended questions such as multiple choices and open-ended questions such as short-answer questions. Although open-ended questions are more powerful in evaluating students' learning, grading on such questions are more time-consuming. In some scenarios such as introductory-level courses in college with hundreds of students, or online courses with an even larger scale of learners, the potentially heavy workload due to the manual grading on open-ended short-answer questions hinders their usage in practice. An automated grading system can provide prompt feedback to a learner, can support large scale learning environment, and further facilitate active and life-long learning. The recent development of Large Language Models (LLMs) has shown their strong general capabilities in many tasks. However, how to use them to automatically provide reliable grading and feedback remains a challenge. We propose SteLLA (\underline{St}ructured Grading System Using \underline{LL}Ms with R\underline{A}G), an automatic grading system that performs a structured grading based on Question Answering (QA) techniques, which is empowered by highly relevant augmented information retrieved from the instructor-provided reference answer and rubric.

%It helps evaluate how well students have learned and guides instructors and students for further learning. Such assessment usually includes problems such as multiple choices, yes or no questions, and open-ended questions, in which grading on open-ended questions is more powerful in evaluating students' learning but meanwhile much more time-consuming. Traditionally, assessment is done manually by an instructor who teaches a course by asking students to answer some questions and grading students' responses. But in some scenarios such as introduction-level courses in college with hundreds of students, or online courses with an even larger scale, manually grading a large number of assignments and exams becomes infeasible or impractical. There has emerged a lot of research in building automated grading systems.

%Although grading is important, without any other feedback, the grade is just a number or letter which contains not much useful information to help students learn further. In this sense, feedback is even more important. Furthermore, according to the research, instant or immediate feedback 

The field of automatic grading and feedback systems has been explored through various domains such as programming \cite{2023.EDM-short-papers.37, 8802114} and mathematics \cite{2023.EDM-short-papers.36, DBLP:journals/jcal/BotelhoBEBH23}, as well as on different types of answers such as essays \cite{mansour-etal-2024-large-language, wang-etal-2022-use} and short answers \cite{10.1162/tacl_a_00236, SUZEN2020726, yoon2023short}. Compared with an essay which is usually long and with multiple paragraphs, a short answer is much shorter and with just a couple of sentences. Grading on short answers is more focused on correctness and does not consider text coherence or writing style as in essay grading. SteLLA is a system designed for automatic short-answer grading (ASAG).

%Questions or problems used in learning assessment and to be graded are normally categorized into two types: one is the type of closed-ended questions such as multiple choices, and the other is the type of open-ended questions such as an essay or short answer questions. Grading on closed-ended questions can be easily automated while grading on open-ended questions is much more challenging. 

%Based on the different types of questions or problems to be graded, this grading task is normally formulated into two categories: one is on close-ended questions such as multiple choices, and the other is on open-ended questions such as an essay or short answer questions. Compared with grading close-ended questions, grading open-ended questions is much more challenging. An essay is usually a long text with multiple paragraphs. Grading on an essay is more focused on characteristics such as coherence, and writing style. In contrast, a short answer is normally composed of a couple of sentences in one or two paragraphs. Automated Short Answer Grading (ASAG) is more focused on the correctness of the answer itself. In this sense, our research falls into the field of ASAG.

There have been many attempts to build automatic grading and feedback systems. Many of them utilize the recent development in Natural Language Processing (NLP). Motivated by the huge progress of LLMs and the needs of instructors and learners, our design uses LLMs as a key component. To ground a general LLM on the specific task of grading, we propose a reference answer and rubric based retrieval augmented generation (R-RAG) approach. Given an instructor-provided reference answer and a rubric, R-RAG extracts highly relevant and structured information from them. It applies question-generation and question-answering techniques to generate a set of evaluation questions and corresponding answers. An LLM performs a structured grading by checking how well a student's response answers these evaluation questions. Eventually, an overall grade, the breakdown grades and feedback are generated to the user.

%approach is to use LLMs and other NLP techniques, e.g., question generation and question answering, to automatically evaluate a student's response to a problem in a structured way. Given a problem, an instructor-provided reference answer, a grading rubric, and a student response, the system will provide not only the overall grade and feedback but also the grades and feedback on all the rubric points.

%Key research questions we ask are: 
%a) If we treat LLMs as a knowledge base, can QA-based approach effectively extract structured information from such a large knowledge base for ASAG task?
%b) In the setting of few-shot learning, can clustering be applied to better select "shots"? 

The contributions of this work are as follows: 
\begin{itemize}
\item We propose an LLM-based ASAG system, SteLLA, that shows substantial agreement with human graders.
\item We present R-RAG which is specifically designed for the grading task. It treats an instructor-provided reference answer and rubric as a knowledge base to extract highly relevant augmented information to ground a general-trained LLM on the grading task.
\item Our system is the first attempt to apply a QA-based structured grading. Compared with the text-similarity-based grading approach, i.e., directly comparing the similarity between the student answer and the reference answer, the QA-based approach provides a tool to induce a deeper semantic understanding of the text in grading. Moreover, It provides not only an overall grade but also decomposed grades and feedback on the knowledge points examined in the problem.
\item We systematically analyze the responses generated by an LLM and show both of its capability and the errors it is prone to make, which provides some insights on how to properly use an LLM in the grading task.
\end{itemize}

The rest of the paper is organized as follows: Section II introduces the background and related work; method and system architecture are explained in Section III; how we collected the data is described in Section IV; Section V presents the experiments and results; the last section, Section VI, gives the conclusion and future work.

\section{Background and Related Work}

\subsection{QA-Based Evaluation}\label{subsec3}
While Question Answering itself is one of the major tasks in NLP, the QA-based approach is novel in applying QA techniques to perform text evaluation for other NLP tasks. This approach has been applied to evaluate the quality of texts in summarization or text compression tasks. Some of the earlier work used the QA evaluation diagram to examine to what extent documents could be summarized while not affecting comprehension on them\cite{10.1287/isre.3.1.17}, to perform human evaluations of summaries\cite{clarke-lapata-2010-discourse}. Along with the progress of question generation techniques, multiple researches have been done on automatically generate questions from the reference summary\cite{chen2018semantic, eyal-etal-2019-question}, from the source document\cite{scialom-etal-2019-answers}, and from the evaluated summary\cite{wang-etal-2020-asking, durmus-etal-2020-feqa} to check fact-based consistency or faithfulness. Extended from previous work, QuestEval combines both recall and precision approaches and shows an improved QA-based metric on evaluating summarization\cite{scialom-etal-2021-questeval}. QestEval is also applied on evaluating text simplification\cite{scialom2021rethinking} and text converted from semi-structured data such as table\cite{rebuffel2021dataquesteval}. To the best of our knowledge, our work is the first attempt to apply QA-based evaluation to the grading and feedback task.

\subsection{Large Language Models (LLMs)}\label{subsec2}
Language Modeling (LM) has been one of the central tasks in NLP. In general, LM is to learn a probability distribution over sequences of tokens by predicting the probabilities of the next or missing token(s). Pre-trained language models such as BERT\cite{devlin2019bert} have shown surprising capability in learning context-aware word representations and achieved high performance in a series of NLP tasks. Since the launching of GPT-3\cite{NEURIPS2020_1457c0d6}, LLMs have attracted a lot of attention. Compared with pre-trained language models, LLMs are scaled with a much larger size of model parameters and training data. They show emerging abilities to solve more complex tasks. ChatGPT (OpenAI 2022), developed upon the GPT-3 (OpenAI 2021) and above series, provides a highly accessible and effective way to use LLMs in a conversational manner and without fine-tuning. This intimates a large number of research and applications. The most recent versions, GPT-4 (OpenAI 2023) and GPT-4O (OpenAI 2024), are multimodel models that accept both text and images as inputs.

Recent LLMs use Transformer\cite{vaswani2023attention} as the backbone architecture of the models. Originally introduced for the machine translation task, the vanilla Transformer is built on an encoder-decoder structure. The encoder and decoder are both a stack of transformer blocks. Through the multi-head attention mechanism, the encoder encodes the input sentence in one language into a latent space of representation; the decoder decodes this representation to autoregressively generate the translated sentence. Different from the vanilla Transformer, the GPT series uses the decoder only.

%Some of the important techniques that are applied in Transformers include a) Self-attention ...... b) Position embedding ...... Different from the vanilla form the Transformer used in the GPT series contains only decoder. 

%\textbf{Prompting} An LLM could be considered a knowledgeable generalist since it is such a large model trained on massive texts. A problem in using it is how to elicit the relevant knowledge when it is used on some specific task, for example, some task in some specific domain such as Biology. Many researches have emerged to study prompting in using a LLM. A prompt is some text which contains a description of the task, and may or may not contain some example(s). 

\subsection{Retrieval-Augmented Generation}
Although LLMs have shown strong general capabilities, there are some key challenges these models are still suffering from, e.g., factual hallucination\cite{cao-etal-2020-factual,raunak-etal-2021-curious,Ji_2023}. Retrieval-Augmented Generation (RAG) \cite{10.5555/3495724.3496517,karpukhin-etal-2020-dense} has been proposed and established to be a technique to alleviate such challenges. It references reliable external knowledge by retrieving relevant information and further enhances the performance of LLMs. Some of the works use the retrieved data as augmented inputs to guide the generation of LLMs \cite{10.5555/3524938.3525306, 10.5555/3495724.3496517}. Others apply this approach in the middle of generation \cite{izacard-grave-2021-leveraging, borgeaud2022improving} or after the generation \cite{khandelwal2020generalization, he2021efficient}. We apply RAG  by using it to retrieve augmented information as inputs. We treat an instructor-provided reference answer as an external knowledge base, extract information that contains the target answer to an evaluation question, and send it together with the student response and the evaluation question to an LLM to perform the assessment.

\begin{figure*}
\centering
\centerline{\includegraphics[width=\textwidth, height=6in]{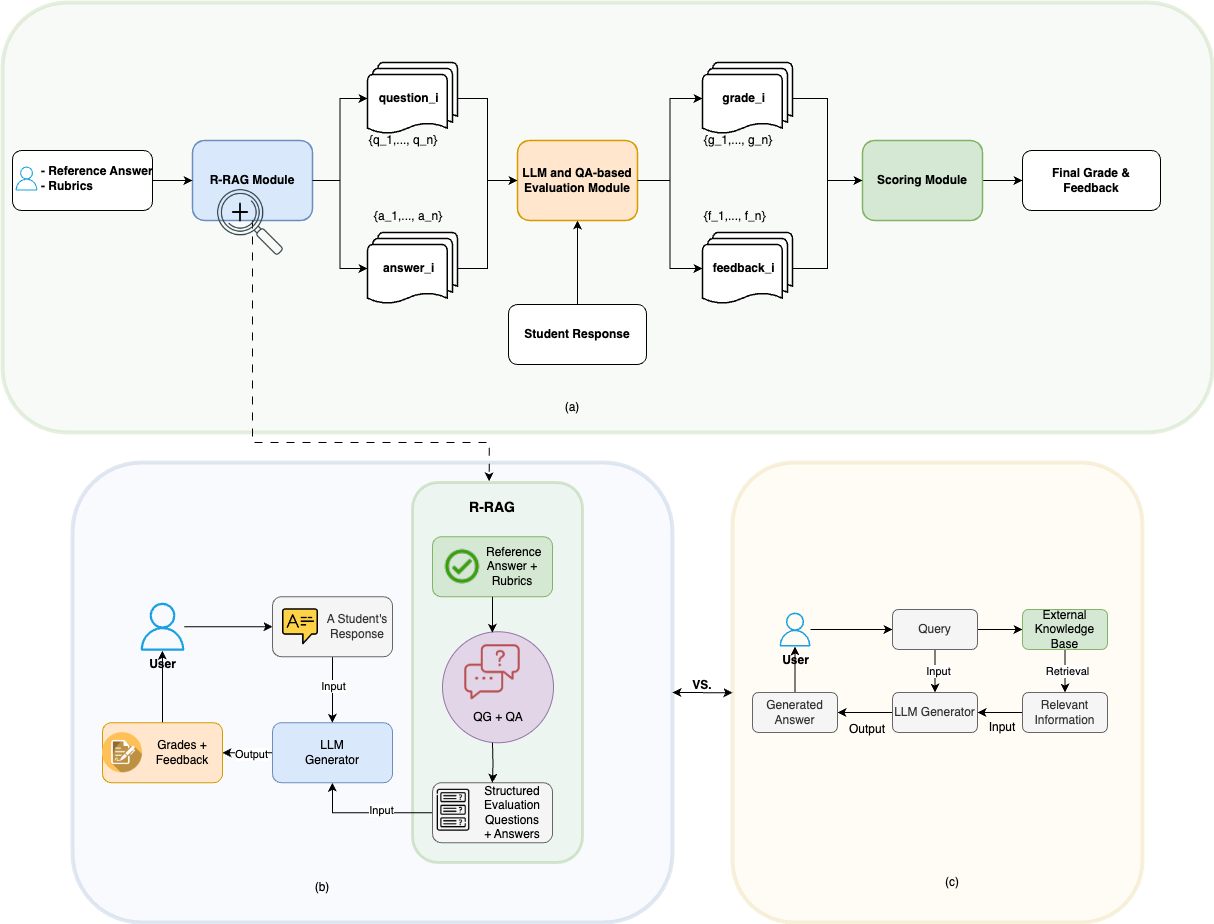}}
\caption{(a) System architecture of SteLLA consisting of i) R-RAG Module which takes the instructor-provided reference answer and rubrics as inputs, generates and extracts a list of evaluation questions with gold answers, and sends it to the LLM; ii) LLM and QA-based Evaluation Module in which an LLM is prompted to perform grading using QA-based evaluation approach; iii) Scoring Module which generates a final grade and feedback. (b) R-RAG approach (c) Typical RAG approach }\label{fig1}
\end{figure*}

\subsection{Automatic Short-Answer Grading}\label{sec3}
The research on the ASAG has a long history. In the earlier days of ASAG research, many traditional methods used rule-based models \cite{Burrows2015TheEA}. For example, the idea of Concept Mapping is more rule-based, which breaks the student answers into several concepts and detects if each concept is present or not \cite{burstein-etal-1996-using, Leacock2003CraterAS, mohler-mihalcea-2009-text}. The approach that uses information retrieval techniques is also more rule-based. It usually checks student answers more by relying on pattern matching through, e.g., regular expressions or parsing trees \cite{Mitchell2002TowardsRC, bachman-etal-2002-reliable}. 

Along with the development of machine learning in NLP, it also has become popular in ASAG systems. Some of them apply clustering methods such as grouping together student responses using LDA clustering to lessen the workload for a human grader \cite{10.1162/tacl_a_00236} or using k-means algorithm based on common word similarity \cite{SUZEN2020726}. Others treat it as a classification problem using, for example, a k-nearest neighbor classifier to detect and diagnose semantic errors in student answers \cite{bailey-meurers-2008-diagnosing}. 

Most recently, interest in Pre-trained Language Models (PLMs) and  LLMs has increased significantly. In accordance, there has been many research on the possible applications of LLMs to the educational field \cite{KASNECI2023102274}. PLMs such as BERT can be pre-trained on domain resources to improve ASAG. \cite{yoon2023short} uses LLM-based one-shot prompting and a text similarity scoring model based on Sentence-BERT \cite{reimers2019sentencebert} to grade short answers. \cite{schneider2023llmbased} evaluated using ChatGPT to perform auto-grading on short text answers, in which they use ChatGPT to directly assess answers by both the educator and the students. They concluded that LLMs currently can be used as a complementary viewpoint but are not ready as an independent tool yet. Our approach is different from the above in the way that we use the instructor-provided reference answer and rubrics as highly relevant external knowledge base, extract structured information in the form of evaluation question-answer pairs, and then ask LLMs to assess to what extent a student's response answers all these evaluation questions.

\begin{figure*}
\centering
\includegraphics[width=0.9\textwidth]{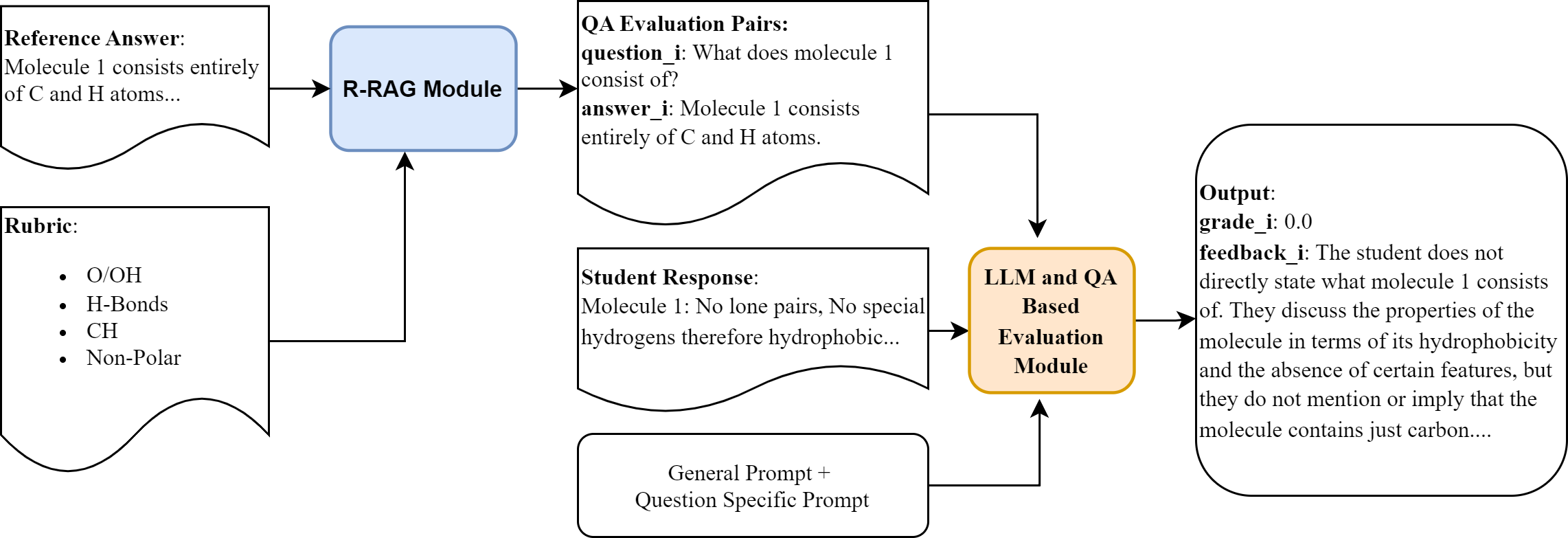}
\caption{An example to show the flow of grading. }\label{fig1}
\end{figure*}

\section{Method and System Architecture}
In this section, we present our approach and the system design. The overall method is to apply the RAG approach to generate structured evaluation questions and corresponding answers from the instructor-provided reference answer and rubrics to a problem. These augmented evaluation question-answer pairs are used to ground an LLM's grading. Together with a student's response and prompts, they are sent to an LLM as inputs. The LLM performs the question-answering task to assess to what extent a student's response answers all these evaluation questions and gives the grades and feedback. The grades of all these questions are eventually consolidated into a final grade. Figure 1 part(a) shows the design of the entire system. A concrete example in Figure 2 illustrates the flow of grading. The prosed system is composed of three key modules: a) the R-RAG module based on the reference answer and rubric, b) the Evaluation module based on LLM and QA, and c) the Scoring module. All the modules are explained in detail in the following.

%\begin{figure*}[htp]
%  \centering
%  \subfigure{\includegraphics[scale=0.5]{rag.png}}
%  \qquad
%  \subfigure{\includegraphics[scale=0.5]{r-rag.png}}
%  \caption{(a) Typical RAG approach (b) Grading task specific RAG approach}
%\end{figure*}

%\textbf{Reference-based RAG Approach} We apply RAG specifically in the setting of the grading task. During the learning assessment, an instructor normally designs a problem and provides a reference response or a grading rubric to it. These are highly reliable which is supposed to be very valuable to ground an LLM to grade a student's response specifically to this problem.

\subsection{R-RAG Module} 
The R-RAG module applies RAG approach based on the reference answer and rubrics. It is specifically designed for the grading task. A typical RAG approach is shown in Figure 1 part (c). Given a query from the user, a retriever, usually using information retrieval techniques, retrieves relevant information from an external knowledge base such as Wikipedia or other reliable datasets. This highly relevant information serves as part of the prompts and guides the LLM to generate specific results for the given query. 

As shown in Figure 1 part (b), the R-RAG module takes the instructor-provided reference answer and rubrics as inputs, generates and extracts a list of evaluation questions with gold answers, and sends it to the LLM. More specifically, given a full-credit reference response \(r\) and a rubric \(b\), each rubric point is marked as a conditioned target answer. A question-generation model will generate a corresponding question for each target answer based on the reference answer. For example, For the rubric point \enquote{C and H}, the corresponding question could be \enquote{What does molecule 1 consist of?} Eventually, this module will generate a set of evaluation questions \(Q=\{q_1,...,q_n\}\) and their gold answers \(L=\{l_1,...,l_n\}\), where \(n\) is the length of the rubric points. Each evaluation question reflects a rubric point. Each gold answer is supported by both the reference response and the rubric. The R-RAG module has some unique designs specifically for the grading task.

\textbf{Highly Relevant Knowledge Base.}
R-RAG treats the instructor-provided reference answers and rubrics as an external knowledge base, which is highly relevant to the grading task that the LLM is going to perform. Normally, the external knowledge that the RAG approach relies on is very large and requires sophisticated techniques to retrieve query-relevant information. Inspired by the traditional learning assessment process in which an instructor usually provides a reference answer and rubrics to facilitate graders in grading, we directly use such available data as external knowledge. They are small and highly relevant to the student's responses that are needed to be graded. This gives the potential to simplify the system and further enhance its usage.

\textbf{Structured Information.}
Due to the nature of external knowledge typically used in RAG which is large, some information retrieval techniques such as ranking are usually used to get the most relevant information. In R-RAG, instead of retrieving ranked relevant information, we aim to extract structured information. This is chosen to perform a structured assessment. To a learner, while it's important to get a correct grade on the answer, it's even more important to understand the knowledge points tested in the problem and how he/she does on each of them. A structured assessment provides more valuable feedback to improve both learning and teaching. Under this consideration, the outputs from the R-RAG are structured following the rubrics, each of which reflects a rubric point. 

\textbf{QA-Based Evaluation.}
When humans grade a student's response to a problem, we do not just compare how similar it is with the reference answer. Instead, for each knowledge point, we ask if the student's response answers it correctly. Inspired by this human grading process, question-answering becomes a natural approach in our automatic grading system. Each bullet point in a rubric is marked as a conditioned answer, a question generation model is applied to generate a question to it based on the reference answer. Meanwhile, a subset of the reference answer which contains the conditioned answer phrase is also extracted for the generated question. They form a question-answer pair. A list of such pairs will be sent as part of inputs to the LLM.

%\begin{figure}[htbp]
%\centering
%\includegraphics[scale=0.9]{rubric.jpg}
%\caption{Example of a reference response and rubric provided by the instructor.}
%\label{fig}
%\end{figure}

\subsection{LLM-based Evaluation Module}
The LLM-based evaluation module takes the outputs from the R-RAG Module, a student's response, and other prompts as inputs. The outputs from this module are a set of numeric grades and detailed feedback to justify its grading.

We apply zero-shot and few-shot learning when prompting the LLM. To better select shots, which are a few task-specific samples provided to an LLM, we use clustering techniques to select learning samples. All students' responses are sent to a sentence encoder such as SBERT \cite{reimers2019sentencebert} to get their embeddings. Then a clustering algorithm such as KMeans is applied to group them into \(k\) clusters. The centroids of all clusters are identified and selected as the few-shots. If a centroid is not a student's response, then find the student's response that is the closest to the centroid.

\subsection {Scoring Module}
The Scoring module takes the set of grades and feedback from the Evaluation module as inputs. Based on the weights of each evaluation question, this module performs the calculation such as weighted sum to generate a final grade of a student's response and a unified feedback. Since the final grade \& feedback and the breakdown grades \& feedback are all valuable, they are all presented to the user as the outputs from the system.

%\subsection{Human Control Over the System}
%We position our system as an assistant which means the human user should still have control over the system. We put this idea with a high level of importance in the design of the system. 

%One place that the user has access to is after query-answer pairs are generated. The user can review each query and its answer, determine their importance and correctness, and make any modifications as needed. It's important to be noted that, this step of review only needs to be done once for one problem. Then the modified version of query-answer pairs remain the same for the grading of all the students' responses.

%The other place that the user can intervene is after the LLM's grading on all students responses are done. There are two considerations here. First, Considering the grade, especially in an exam, affects a student's academic performance significantly, false negatives have worse results than false positives. This leads to the idea to give the user the chance to review all the negative grading results. Second, the user can add some personal criteria. For example, the user can set different weight for each query-answer pair which will lead to different final grades.

\section{Data}
In this section, we report the data collected for this study. We first describe the data source, then explain how we redact the data to protect students' privacy, and lastly present statistics of the data.

\subsection{Data Source}
The data used in this study are collected from an undergraduate-level introductory Biology course in the semester of Fall 2018 at a public university in the United States. The data are student's answers to a problem from an exam. We will make the dataset public after publication. As shown in Figure 3, in part (a) of this problem, students are provided with 3 images of different molecules and asked to rank them in the order from the most hydrophobic to the most hydrophilic. In part (b) of the problem, students are asked to briefly explain their choices in part a. Their short answers in part (b) are the data collected for this study.

  \begin{figure*}
  \centering
  \includegraphics[width=0.9\textwidth, height=3in]{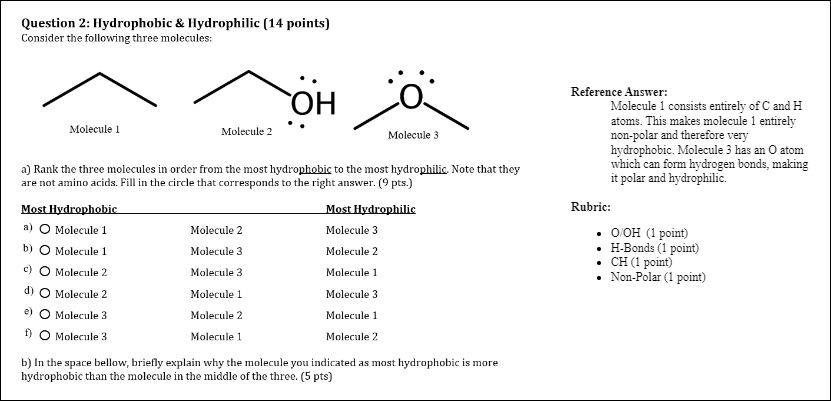}
  \caption{The problem, the reference answer, and the rubric in the dataset.}\label{fig1}
  \end{figure*}

\subsection{Privacy Protection}
We take our responsibility to protect students' privacy seriously. The data used in this study are all under the approval of the Institutional Review Board (IRB) at the school where the data are collected. We redact the data to make them de-identified through the following pre-processing: a) Removing student names and using file names as index instead; b) Removing any information in the answers that can be linked to any specific individual.

\subsection{Labeling Process}
Two undergraduate Research Assistants, who had taken the same Biology course before and understood the course materials well, did the labeling as human graders. The entire labeling is an iterated process, two graders working first respectively to give only a final grade to each student's answer, then adding grades to all rubric points, and in the end consolidating two graders' labels into an agreed version. For the selected few-shot samples, the human graders also give the text feedback to justify their grading. This process lasted about two semesters.

The two human graders are first trained by the instructor on how to do grading specifically for assignments or exams for this course. Then they label the data in two steps. In step one, they do the labeling respectively. For each evaluation question on a problem, they check to what extent a student's response answers the question. If it answers the question completely correctly, then label it with 1; otherwise, label it with 0. We do not consider partial credits since the evaluation has been decomposed into a set of questions, each of which is focused on one knowledge point. We original start with only the one final grade for each answer. Along with the development of the approach, the graders are instructed to add labels to all the rubric points for a problem. Then in step two, under the guidance of the instructor, the two human graders identify all the labels that they do not agree with each other, have a discussion, and come across the labels they all agree. Eventually, this process gives us the ground-truth labels for evaluation.

\subsection{Characteristics and Statistics}
The collected data contain a total of 176 samples. Due to one empty entry, the number of valid samples is 175. The average length of a student's answer is around 39 words. Each answer, which is a paragraph, contains around 2 sentences on average. This is consistent with the normal description of short answers such as the length is \enquote{phrases to three to four sentences} or \enquote{a few words to approximately 100 words} \cite{sukkarieh-stoyanchev-2009-automating}.

%\begin{table}[htbp]
%\caption{Statistics of the data set}
%\begin{center}
%\begin{tabular}{|c|c|c|c|}
%\hline
%\textbf{Table}&\multicolumn{3}{|c|}{\textbf{----}} \\
%\cline{2-4} 
%\textbf{Head} & \textbf{\textit{Total}}& \textbf{\textit{Ave. Words}}& \textbf{\textit{Avg. Sentences}} \\
%\hline
%copy& 175&39 & 1.8 \\
%\hline

%\end{tabular}
%\label{tab1}
%\end{center}
%\end{table}

To facilitate the human graders, the instructor provides one reference answer and a grading rubric which contains 4 key rubric points such as O/OH and H-Bonds. The score of each rubric point is 1. This leads to 4 points total as the full score for the problem. Originally part b in the exam is 5 points. The instructor adjusted it to be 4 points based on the rubric. Accordingly, the score on each rubric point is binary (0/1) and the score of each student is an integer value in the range of 0-4 inclusive. The following are the reference answer and a sample student answer:

\begin{quote}
    Reference answer: \textit{Molecule 1 consists entirely of C and H atoms. This makes molecule 1 entirely non-polar and therefore very hydrophobic. Molecule 3 has an O atom which can form hydrogen bonds, making it polar and hydrophilic.} \\

    \noindent
    Sample student answer: \textit{Molecule 1: No lone pairs, No special hydrogens therefore hydrophobic. Molecule 2: Two lone pairs, has special hydrogen therefore more hydrophilic than molecule 1}
\end{quote}

\section{Experiments and Results}
In this section, we describe our experiment settings and report the experiment results.

\subsection{Experiment Settings}
In the R-RAG module, the instructor-provided reference answer and the rubric are both supplied. Answer-conditioned question generation is applied to the reference answer, in which one rubric point is set as a conditioned answer to generate one question. To make sure the generated questions are of high quality then we can have a more consistent and solid evaluation of LLM’s performance, we manually generate three questions for each rubric point based on the reference answer. The course instructor reviews these questions and selects the best one out of the three. In the Evaluation Module, the system calls GPT4 API (the version of GPT4-Turbo-Preview). When prompting GPT4, we design general instruction and question-specific instruction. The general instruction is to specify the role, task, detailed instruction, and constraints on how to grade such as the grade scale, criteria of each grade, etc. The question-specific instruction is to address a grader's personal criteria. For example, in the evaluation question "What does molecule 1 consist of?", although the reference answer expects a student's answer to contain the information that molecule 1 consists of C and H atoms, the course instructor thinks if a student's answer only mentions C (carbon) atom, it's also considered as being correct. This personal criteria is addressed in the question-specific instruction. We apply few-shot learning in which the 4-shot gives us the best performance. To select samples, we perform random selection. Selected samples are excluded from evaluation. The following shows an example of instructions in the prompt:

\begin{quote}
    General instruction: \textit{You are the instructor of a college-level Introductory Biology course. You are going to grade the exam for this course. Your grading should be based on the question asked, the full-credit answer, the student's answer, and nothing else. Give the binary score 1 or 0, in which 1 means the student's answer is correct and 0 means the student's answer is incorrect or does not answer the question, and justify your grading.} \\

    \noindent
    Question-specific instruction: \textit{As long as the answer mentions or implies that the molecule contains just carbon, it should be considered as being correct and graded as 1.}
\end{quote}

\subsection{Evaluation Results}
SteLLA essentially takes the role of a grader. Thus We evaluate the results by calculating the agreement with the human grader's grading, which is commonly used in grading evaluation. Because this work is pioneering in applying QA-based evaluation on ASAG task, on a newly collected real-world dataset, and this field is relatively new, we weren't able to find highly related models or systems to compare with. As explained in the labeling process section, under the instructor's supervision, the two human graders discussed the difference in the grades they assigned to the same questions, reached an agreement, and reassigned the agreed grades to those questions as the ground-truth labels. We compare the agreement between the results from our system and the ground-truth labels and report both Cohen's Kappa coefficient ($\kappa$) \cite{cohen1960} and Raw Agreement (Accuracy).

\textbf{Agreement Results.} As shown in Table I, Cohen's Kappa coefficient value between the human grader and the ground-truth labels reaches 0.8315 which is normally accepted as a near-perfect agreement. Although our system still does not reach human performance, it achieves a substantial agreement with the ground-truth labels by \(\kappa = 0.6720\). As for the raw agreement, it's about 8\% lower than the human grader. These results show that our system is promising in automatic grading while maintaining high accuracy.

\begin{table}[htbp]
\centering
\caption{Agreement results between the system and labels}\label{tab1}%
\begin{tabular}{|c|c|c|}
%\toprule
\hline
    & Cohen's Kappa  & Raw Agreement  \rule[-2ex]{0pt}{6ex}\\
%\midrule
\hline
Human       & 0.8315   & 0.9157  \rule[-1ex]{0pt}{4ex}\\
\hline
Our System    & 0.6720   & 0.8358  \rule[-1ex]{0pt}{4ex}\\
%\botrule
\hline
\end{tabular}
\end{table}

\begin{table*}[]
\renewcommand*{\arraystretch}{1.1}
\centering
\caption{Example gradings and feedback}
\begin{tabular*}{\textwidth}{|p{0.9cm}|p{15.5cm}|p{0.5cm}|} 
\hline
Question & GPT4's Grading & Label \\
\hline
\multicolumn{3}{|c|}{} \\[-8pt]
\hline
\multicolumn{3}{|p{15cm}|}{Q1: Does molecule 3 have Oxygen (O) atom or OH Group? }\\
\multicolumn{3}{|p{16cm}|}{Q2: Can molecule 3 form H-Bonds or Hydrogen Bonds?}\\
\multicolumn{3}{|p{16cm}|}{Q3: What does molecule 1 consist of?}\\
\multicolumn{3}{|p{16cm}|}{Q4: Is molecule 1 entirely non-polar?}\\
\hline
\multicolumn{3}{|c|}{} \\[-8pt]
\hline
\multicolumn{3}{|p{15cm}|}{Student response 9328795: Molecule 1 is the most hydrophobic because it is a carbon chain tho does not have the ability to create hydrogen or ionic bonds. Whereas Molecule 3 is more hydrophilic then molecule 1 due to its ability to form hydrogen bonds because of the lone pairs on the oxygen atom. }\\
\hline
Q1 & The student’s score is 1. The student answers the question correctly. The student states, \enquote{Molecule 3 is more hydrophilic then molecule 1 due to its ability to form hydrogen bonds because of the lone pairs on the oxygen atom.} \textit{\textcolor{green}{\textbf{This indicates that molecule 3 has an Oxygen atom}}}, which is consistent with the full-credit answer. Accordingly, the student answers the question correctly. & 1\\ 
\hline
Q2 & The student’s score is 1. The student answers the question correctly. The student's answer states, \enquote{Molecule 3 is more hydrophilic then molecule 1 due to its ability to form hydrogen bonds because of the lone pairs on the oxygen atom,} \textit{\textcolor{green}{\textbf{indicating that molecule 3 can form hydrogen bonds}}}. This aligns with the full-credit answer which states \enquote{Molecule 3 has an O atom which can form hydrogen bonds}. Accordingly, the student answers the question correctly. & 1 \\ 
\hline
Q3 & The student’s score is 1. The student answers the question correctly. The student states, \enquote{Molecule 1 is the most hydrophobic because it is a carbon chain,} \textit{\textcolor{green}{\textbf{implying it consists of Carbon}}}. Moreover, the student says that molecule 1 \enquote{does not have the ability to create hydrogen or ionic bonds,} which \textit{\textcolor{green}{\textbf{indirectly suggests the presence of Hydrogen}}}. Both of these components are consistent with the full-credit answer, "Molecule 1 consists entirely of C and H atoms." Accordingly, the student answers the question correctly. & 1 \\ 
\hline
Q4 & The student’s score is 0. The student's answer is not relevant to the question. Even though the student discusses the hydrophobic nature of molecule 1, they\textit{ \textcolor{green}{\textbf{do not answer directly if molecule 1 is entirely non-polar}}}. Accordingly, the student's answer is not relevant to this question. & 0 \\
\hline
\multicolumn{3}{|c|}{} \\[-6pt]
\hline
\multicolumn{3}{|p{15cm}|}{Student response 9328790: Molecule 1 is most hydrophobic because it is all carbons and it can't make hydrogen bonds. But Molecule \#2 has an OH at the end, allowing it to create h-bonds.}\\
\hline
Q1 & The student’s score is 0. The student's answer is not relevant to the question. \textit{\textcolor{blue}{\textbf{The student's answer does not include any information about molecule 3. Instead, the answer discusses molecules 1 and 2.}}} Accordingly, the student answer is not relevant to the question. & 1 \\
\hline
Q2 & The student’s score is 0. The student does not answer the question. \textit{\textcolor{blue}{\textbf{The student provides information about molecules 1 and 2, but does not mention whether or not molecule 3 can form hydrogen bonds.}}} Accordingly, the student does not answer the question. & 1 \\
\hline
Q3 & The student’s score is 1. The student answers the question correctly. The student states, \enquote{Molecule 1 is most hydrophobic because it is all carbons and it can't make hydrogen bonds.} \textit{\textcolor{green}{\textbf{This implies that molecule 1 consists of carbon atoms and does not contain elements like oxygen or nitrogen which can form hydrogen bonds.}}} This is consistent with the full-credit answer \enquote{Molecule 1 consists entirely of C and H atoms.} Accordingly, the student answers the question correctly. & 1 \\
\hline
Q4 & The student’s score is 0. The student's answer does not directly address the question asked. \textit{\textcolor{green}{{While the student correctly identifies molecule 1 as being hydrophobic and incapable of forming hydrogen bonds, they do not explicitly state that this makes molecule 1 entirely non-polar.}}} The student also brings in comparison with other molecules (Molecule \#2), which is not relevant to the specific question asked. Therefore, the answer is incorrect. & 0\\
\hline
\multicolumn{3}{|c|}{} \\[-8pt]
\hline
\multicolumn{3}{|p{15cm}|}{Student response 9328809: Molecule 1 does not have donor or acceptor. It is probably a covalent bond, which is the strongest bond. Molecule 2 has a -OH bond, which makes it partly hydrophilic, but more hydrophobic. Molecule 3 only has an O atom, which is an acceptor, but cannot form H-bonds.}\\
\hline
 Q1 & The student’s score is 1. The student answers the question correctly. The student states, \enquote{Molecule 3 only has an O atom, which is an acceptor, but cannot form H-bonds.} \textit{\textcolor{green}{\textbf{This indicates that molecule 3 has an O atom}}}, which aligns with the full-credit answer, "Molecule 3 has an O atom". Accordingly, the student answers the question correctly. & 1\\
\hline
Q2 & The student’s score is 0. The student answers the question incorrectly. The student's answer \enquote{Molecule 3 only has an O atom, which is an acceptor, but cannot form H-bonds} contradicts the full-credit answer \enquote{Molecule 3 has an O atom which can form hydrogen bonds}. \textit{\textcolor{green}{\textbf{An oxygen atom in a molecule can be a hydrogen bond acceptor, meaning that it can form a hydrogen bond. Therefore, the student's understanding of the concept is incorrect.}}} & 0 \\
\hline
Q3 & The student’s score is 0. The student's answer is not relevant to the question. \textit{\textcolor{green}{\textbf{The student mentions various properties of molecule 1, but does not state what molecule 1 consists of in terms of atoms.}}} Accordingly, the student's answer is not relevant to the question. & 0 \\
\hline
Q4 & The student’s score is 1. The student answers the question correctly. \textit{\textcolor{red}{\textbf{The student's response includes \enquote{Molecule 1 does not have donor or acceptor}, which can be interpreted as suggesting that molecule 1 is non-polar}}}, aligning with the full-credit answer. Thus, the student correctly answers the question. & 0\\
\hline

\end{tabular*}
\end{table*}

\textbf{Human Evaluation on Feedback.} In order to further understand the generated text from LLM which is to justify the grading, we did a human evaluation of all the justifications generated by GPT4. The two human graders are instructed to do the evaluation. In human evaluation, the question we ask is how relevant the justification generated by GPT4 is to support its grading. In other words, if the grade assigned by GPT4 is correct or incorrect, does the justification support this grading? The data we use for human evaluation is from a 6-shot learning experiment setting which leaves a total of 169 samples for evaluation. Since 4 evaluation questions are generated for the problem, there is a total of 676 GPT4 responses to be evaluated. Very surprisingly, only 1 response is evaluated to be irrelevant to the numeric grade. Even when the grading of GPT4 is incorrect, it's usually still based on the relevant facts but with too much or not enough inference which will be shown in the sample results analysis in the following. This shows that GPT4 does do the grading based on the relevant facts which increases the confidence in using an application based on it.

\subsection{Sample Grading and Feedback Analysis}
In Table II, we list three sample students' responses and GPT4 grading results to the evaluation questions. We have several findings about using GPT4 to do grading as:

\begin{itemize}
\item GPT4 is good at identifying relevant facts or statements. For example, in Q1 and Q2 to the student response 9328795, GPT4 is able to identify that molecule 3 has an Oxygen atom and can form hydrogen bonds even though the two phrases are a bit far from each other in the original text answer. In student response 9328809, GPT4 identifies question-related information that molecule 3 has an O atom and it cannot form H-bonds and then grades the student response on Q1 is correct and on Q3 is incorrect.
\item GPT4 can be tolerant of some typos in the input. For example, in student response 9328795, there are typos or errors such as tho and then. But they do not affect GPT4's understanding of the response text.
\item GPT4 sometimes can infer the meaning of the text properly, while sometimes infers too much implication from the given text. For example, in Q3 to the student response 9328790, based on 
\enquote{Molecule 1 is most hydrophobic because it is all carbons and it can't make hydrogen bonds.}, GPT4 properly infers that the student implies molecule 1 consists of carbon atoms and does not contain elements like oxygen or nitrogen which can form hydrogen bonds, and further grades it as being correct on this evaluation question. While in Q4 to the student response 9328809, GPT4 interprets the student's statement \enquote{Molecule 1 does not have donor or acceptor} as suggesting that molecule 1 is non-polar and grades it as being correct which is actually incorrect. In this example, GPT4's interpretation might be true in general. However, it infers too much from the student's response in this specific problem, in which the instructor tries to test the concept of non-polar. We notice this is a type of error that GPT4 is prone to make in this grading task. This error type shows that, since LLM such as GPT is trained on massive data which is expected to have learned a large amount of general knowledge, how to ground it to some specific task and some specific domain is a big challenge. Our methods of R-RAG and structured evaluation provide an approach to address this issue. We also experimented with prompting engineering to set some constraints, such as defining the role to be a college-level Biology instructor and explicitly asking GPT4 to do the grading based only on the student's response, the evaluation question, the reference answer to the question, and nothing else. However, we find it's still hard to eliminate such error types by refining the prompts only.
\item Error cases of Q1 and Q2 in the student response 9328790 show the complexity of the grading task. Due to the student not giving any statements about molecule 3, GPT4 grades the response to be incorrect on these two questions which are both about molecule 3. However, the human grader is more focused on the concept that the most hydrophilic molecule has an OH which makes it able to form H-Bonds. Based on this, although the student discusses molecule 2 instead of molecule 3, the response shows he/she indeed understands the concept correctly. Accordingly, human graders give the student full credit on these two questions. During the human evaluation process, the course instructor and two human graders all agree that, in such cases, GPT4 does the job properly based on the instructions it's given. The challenge lies not only in how to make an LLM understand the abstract concept behind the text, but also in how to formulate what is examined in a problem in the learning process itself.

\end{itemize}

\subsection{Ablation Study}
We did the following ablation studies to show the effect of some parameters and settings. Due to the time and cost constraints, the following experiments were done using GPT-4.

\textbf{Effect of Clustering.}
As shown in Figure 4, applying a clustering algorithm to select samples for few-shot learning consistently improves Cohen's Kappa coefficient compared with that without using clustering, e.g., about 0.2 increments in the \(\kappa\) value under one-shot. This supports the effectiveness of the clustering approach in selecting learning samples that are expected to better represent the distribution of the data, and further empower the capability of the LLM such as GPT4 on this specific dataset and task.

\textbf{Number of Shots.} We experimented with different shot numbers. The Cohen's Kappa coefficient values in Figure 4 show that a few learning samples can significantly improve the performance of a general LLM on a specific task such as grading. Under the setting with clustering, the 4-shot gives the best result which is significantly higher than the 3-shot while slightly higher than the 5-shot. Under the setting without clustering, the performance under the 6-shot is significantly better than the 1-shot, while the 10-shot does not show much further improvement compared with the 6-shot. This is consistent with the common understanding that the few-shot in-text learning can guide a general LLM toward a specific task such as grading in this experiment. Meanwhile, the effect declines when reaching a reasonable shot number. 
%One exception we observe is that, under the setting without clustering, 1-shot learning leads to a significant decrease in performance compared with that under 0-shot. 

%The 4-shot as shown in Figure 5 gives us the best performance which is significantly higher than 3-shot while slightly higher than 5-shot. From 1-shot to 4-shot, the Cohen's Kappa coefficient and the accuracy both consistently increase. The only exception, which is surprisingly interesting, is with 0-shot. Based on the results from 0-shot which means no learning samples are provided to GPT4 at all, it can already achieve better performance than 3-shot setting. This might indicate LLM such as GPT4 indeed learns a significant amount of knowledge during its pre-training process which gives it some capability to work on any task. When some learning samples are provided for some specific task, it takes a reasonable number of samples to direct the learned general knowledge toward the task.

\begin{figure}[htp]
    \centering
    \begin{tikzpicture}
        \begin{axis}
            [width=3.4in,height=7cm,xlabel={N-shot},
            ylabel={Agreement}, grid=major, domain=1:35, xmin=0, xmax=10,
            ymin=0, ymax=1, xtick={1,2,...,10}, ytick={0,0.2,...,1},
            xlabel style={black!70}, ylabel style={black!70},
            tick label style={black!70}, tick style={black!50},
            axis line style={black!50}, legend style={draw=black!50},
            legend pos=north east, samples=10, grid style=solid]
            \addplot[c1, very thick, dashed, mark=triangle*,mark size=3,mark options={c1, very thick, solid, fill=white}] 
            coordinates {(1,0.3192)(2,0.3894)(3,0.4312)(4,0.4875)(5,0.4807)}; 
            \addlegendentry{$\kappa$ with clustering};
            %\addplot[c2, very thick, dotted, mark=square*, mark size=3, mark options={c2, very thick, solid, fill=white}] 
            %coordinates {(0,0.5759)(1,0.6621)(2,0.6966)(3,0.7172)(4,0.7448)(5,0.7414)}; \addlegendentry{AccuracywithclusteringAccuracy with clustering};
            \addplot[c2, very thick, dotted, mark=diamond*,mark size=3,mark options={c2, very thick, solid, fill=white}] 
            coordinates {(1,0.118)(6,0.385)(10,0.391)}; 
            \addlegendentry{$\kappa$ without clustering};
        \end{axis}
    \end{tikzpicture}
    \vspace*{-2mm}
    \caption{Effect of shot number.}
\end{figure}
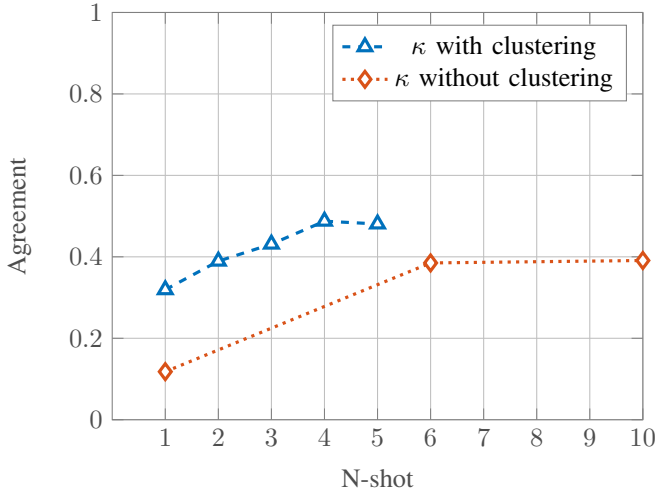

\section{Conclusion and Future Work}
We propose SteLLA, an automatic short-answer grading system that uses RAG techniques based on the instructor-provided reference answer and rubric to facilitate an LLM performing structured question-answering-based assessment of student responses. Experiments on a real-world dataset show that our system is able to achieve substantial agreement with the human graders. It can also provide analytical grades and feedback on knowledge points examined in the problem. In the future, one direction of the work could be on generating structured evaluation question-answer pairs in the context of missing rubrics, i.e., only the reference answer available. Another direction could be to add human-interactive components to increase the system's adaptability in personalization.

\section*{Acknowledgment}
This paragraph is omitted due to the blind review policy.

\end{document}